\documentclass{article}

 \usepackage[preprint]{neurips_2025}

\usepackage[utf8]{inputenc} 
\usepackage[T1]{fontenc}    
\usepackage{hyperref}       
\usepackage{url}            
\usepackage{booktabs}       
\usepackage{amsfonts}       
\usepackage{nicefrac}       
\usepackage{microtype}      
\usepackage{xcolor}         
\usepackage{graphicx}

\title{Human-Machine Ritual: Synergic Performance through Real-Time Motion Recognition}
\author{%
  Zhuodi Cai\thanks{Equal contribution}\quad\quad
  Ziyu Xu\footnotemark[1]\quad\quad 
  Juan Pampin\thanks{Corresponding author} \\
  Department of Digital Arts and Experimental Media (DXARTS), University of Washington \\
  Seattle, WA, USA \\
}

\begin{document}

\maketitle

\begin{abstract}
We introduce a lightweight, real-time motion recognition system that enables synergic human-machine performance through wearable IMU sensor data, MiniRocket time-series classification, and responsive multimedia control. By mapping dancer-specific movement to sound through somatic memory and association, we propose an alternative approach to human-machine collaboration, one that preserves the expressive depth of the performing body while leveraging machine learning for attentive observation and responsiveness. We demonstrate that this human-centered design reliably supports high accuracy classification (<50 ms latency), offering a replicable framework to integrate dance-literate machines into creative, educational, and live performance contexts.
\end{abstract}

\section{Introduction}
Humanity lies in the body, in its sensations, memories, and intuitions. In the contemporary era of new media arts, as artificial intelligence (AI) reshapes the creative landscape, we ask not how well machines can imitate human performance, but how they might respectfully observe and respond to it. In contrast to systems trained on large-scale datasets and generic genre labels, our project centers the dancer’s body as both archive and oracle: a source of deeply personal movement emerging from memory, feeling, and improvisation.

We propose a new ritual of co-performance, one that honors the unique strength of human and machine. Rather than designing a generative system that tries to emulate human creativity, we develop a dancer-specific feedback loop using wearable Inertial Measurement Unit (IMU) sensors and a ridge classifier (with Minimally Random Convolutional Kernel Transform, in short, MiniRocket \cite{minirocket}). This allows the system to recognize patterns of motion in real time and respond sonically, not by inventing, but by recalling sound paths already meaningful to the dancer. The machine does not create, but remembers with the human.

In this collaboration, the data archive, the role of the machine, and the procedure of the exchange are all tailored by the artists. This framework favors attunement over automation, drawing from somatic philosophies that emphasize presence above control.

\section{Background}

Across cultures and histories, human communities have shared the experience of dancing and singing in ceremonies and rituals. These practices connect people, transmit knowledge, and create space for mourning, celebration, expression, and healing. Performance as ritual strengthens interpersonal bonds while allowing cultures to evolve over time.

The significance of performance lies not just in its social function. It is rooted in the body itself.  As Prentis Hemphill writes in Chapter 3, \textit{Feeling and the Body}, of \textit{What It Takes to Heal}, the intrapersonal practice of embodiment allows us to “sync our bodies with one another’s and feel the presence of what we share that lies beyond words” \cite{hemphill2025}. Through the rituals of meditation and somatics, we connect our minds to our physical bodies, to the body of the community around us, and to the humanity that we feel before it is spoken. Originating from the Greek word “soma” (“the living body in its wholeness”), the practice of somatics sees the mind, body, spirit, and the environment where they coexist as an inclusive whole \cite{brodie2012dance}. Through the lens of somatics, dance, music, and performance thus become sites of embodied inquiry, where knowing and feeling are inseparable.

As technologies progressively enter these spaces, machines operating through predefined algorithms rather than lived intelligence are gradually able to imitate humans better and even surpass certain limitations. This growing capacity for imitation is evident in the performing arts, particularly from the 18th to the 21st century. For example, the Jaquet-Droz family’s \textit{Musician} automaton \cite{jaquetdroz} plays music without ever forgetting the score. Reuge’s spinning ballerinas in music boxes \cite{reugeballerina1950s} can be replicated without the ethical concerns of human cloning. Disney’s Audio-Animatronics \cite{disneyanimatronics2024} perform on stage tirelessly. Boston Dynamics’ humanoid robots \cite{bostondynamics2020} dance with remarkable strength and endurance. Suno \cite{sunoai2025} generates commercially viable songs using less time and money than human creators.

While emerging AI may be designed to mimic human appearance, intelligence, and creativity, such imitations risk obscuring the profound differences between human and machine ways of knowing. Human knowledge is not abstracted from the body, but lived through it: a somatic, embodied knowledge that is sensory, emotional, and deeply entangled with the uncertainties and ambiguities of the physical world. This experience is not only rich and unique to each individual but also inseparable from the interconnected wholeness of mind, body, spirit, and their environment. Dance, as a great example, reveals this irreplaceable depth of human embodied knowledge, a dimension of being that no algorithmic entity can reproduce.

\section{Related Work}

\subsection{Human-AI Co-Performance}

In the performing arts, one way that people connect dance and music is through sensorimotor coupling. According to cognitive psychologist Wolfgang Prinz’s common-coding approach \cite{Prinz01061997}, perception and planned actions share a representational domain that can mutually influence or activate each other, enabling a bidirectional link between listening and moving. 

Such a link inspires some AI studies of dance-music association. The EDGE music-to-dance model \cite{EDGE2023} uses a transformer-based diffusion model trained on the AIST++ \cite{AIST++2021} genre-labeled dataset. MusicGen's \cite{musicgen} output is guided by a dance-to-music system \cite{dancetomusic2024} based on genre and video keypoints. Reimagining Dance \cite{vechtomova2025reimagining} describes a system where dancers dynamically control music in real time using the AIST \cite{AIST2019} dataset labeled by genre, making the dancer a co-composer of the soundscape. Although technically functional, by imposing predefined categories and relying on AI’s internal genre logic and fixed definitions of dance and music, these pipelines limit interpretability and omit the artist’s holistic, embodied engagement. Our work, instead, aims to preserve the rich associations that emerge through lived and felt experience by valuing the dancer’s own remembered movement-sound connections, which serve as a bridge for human-AI dance-music synergic performance (see Section \ref{sec:4.2}).

Virtual dance partners driven by AI have been developed to perform with human dancers in real time. Georgia Tech’s LuminAI introduces an improvisational AI dance partner \cite{luminai2017}, and MIT Media Lab’s recent Human-AI Co-Dancing project \cite{mitcodance2024} has explored real-time collaboration between human dancers and virtual agents. These systems often involve AI, generating dance moves or interacting visually, whereas our system keeps the human as the sole mover and the AI as an attentive “stage manager” that controls other media of the performance, such as lighting, sound, even sets and props, for the dance. 

\subsection{Personalized AI Tools for Dance}
Personalized AI tools for dance are still uncommon, but Fiebrink’s Wekinator software pioneered real-time, performer-trained gesture recognition for music control. This tool allows artists to map bodily gestures to sound through an interactive, on-the-fly learning paradigm. The entire procedure, from model training to evaluation, “happens on the order of seconds” \cite{Wekinator_fiebrink_2009_1177513}.

Our work extends this paradigm by adopting a state-of-the-art time-series method and anchoring the mapping in the dancer’s own memories. Differing from Wekinator, which uses discrete gesture input, our method processes continuous motion, aligning more closely with the nature of dance. Moreover, our implementation supports real-time performance with latency typically under 0.1 seconds (see Section \ref{sec:4.3} for details). 

\section{Concept and Methodology}
\subsection{A New Ritual}
The question of how human and machine intergrow remains open. In \textit{A Cyborg Manifesto}, Donna Haraway argues that we are already entangled with our technologies, and that the line between human and machine is less a boundary than a site of possibility \cite{Haraway1991}. Her framework invites new rituals in which bodies and computational systems create shared experiences. We advocate one such ritual: a performance that integrates machine learning while carrying forward emotional depth and creative expression.

This project is choreographed, composed, and engineered in search of self-compassion. Inspired by the image of a lake in somatic meditation, we use music, dance, visual art, and interactive systems to embody quiet strength. A still surface, and beneath it, a restrained, surging undercurrent.

Human artist and machine perception collaborate in stages. The artist creates, the machine learns to perceive this specific art, and the system becomes a bridge for multimedia dialogue. As the dancer moves, the system senses, interprets, and responds through sound and projection. The body leads as the site of somatic expression, while the machine acts as an active participant, shaping space through deep listening \cite{deeplistening2005oliveros}.

Echoing Haraway’s vision for hybrid practices that neither reject technology nor erase human specificity, we put forward a particular combination of somatics, personal memory mapping, along with MiniRocket, that configures into an intimate ritual for human-machine synergic performance. This explores how humanity and performance might evolve in the presence of machine collaborators. In our work, the lake becomes both metaphor and structure. A reflective surface where human and machine meet in presence, mindfulness, and harmony.

\subsection{Embodied Memory and Sound Mapping in Dance}
\label{sec:4.2}

We turn to interactive machine learning through a human-centered lens. The dance-to-music mappings are neither random nor AI-determined. They grow from the dancer’s soma. Instead of predefining gestures or using generic labels, the dancer improvises movements in response to personally evocative sounds, encoding her own memories and imagery into the system. The machine learns to remember these pairings and later triggers the sounds when it detects the motions. The sounds have meaning to the dancer, and her movements are naturally inspired by those sounds, creating a tight feedback loop of meaning. What differentiates our work from prior related works is that co-performance here is realized via recollection instead of generation.

Composing music involves more than triggering isolated samples. To organize sound meaningfully, we draw on the dancer’s verbal descriptions of the memories and imagery evoked by each sample. For example, hearing motion 1’s sound, the dancer says, “I immediately see an image of a metal or glass windchime... So I wiggle my arms and my body, twist my knees as I step, feel the air as I move...” For motion 2, she describes, “It puts me in a tunnel... commuting to work in a subway... the rush, the boredom, the repetition, the life we waste.” These descriptions assist in creating narratives through music in performances.

\subsection{IMU Based Motion Data Collection and Recognition with MiniRocket}
\label{sec:4.3}

MiniRocket \cite{minirocket} is a recently introduced method known for its efficiency in time-series classification. It is not yet commonly seen in interactive art contexts. Many prior interactive dance systems used either simpler heuristics or general-purpose classifiers (SVMs, neural networks via Wekinator \cite{Wekinator_fiebrink_2009_1177513}, etc.) rather than specialized time-series methods. 

We utilize wearable IMU sensors to allow for lightweight, affordable, wireless, three-dimensional, high-frequency movement capture. MiniRocket’s accuracy in combination with IMU’s bespoke sensing gives our dancer confidence that her subtle motions can be reliably witnessed. By deploying IMU based MiniRocket for creative AI, a more immediate, embodied interaction becomes possible.

\begin{figure}[htbp]
  \centering
  \includegraphics[width=\textwidth]{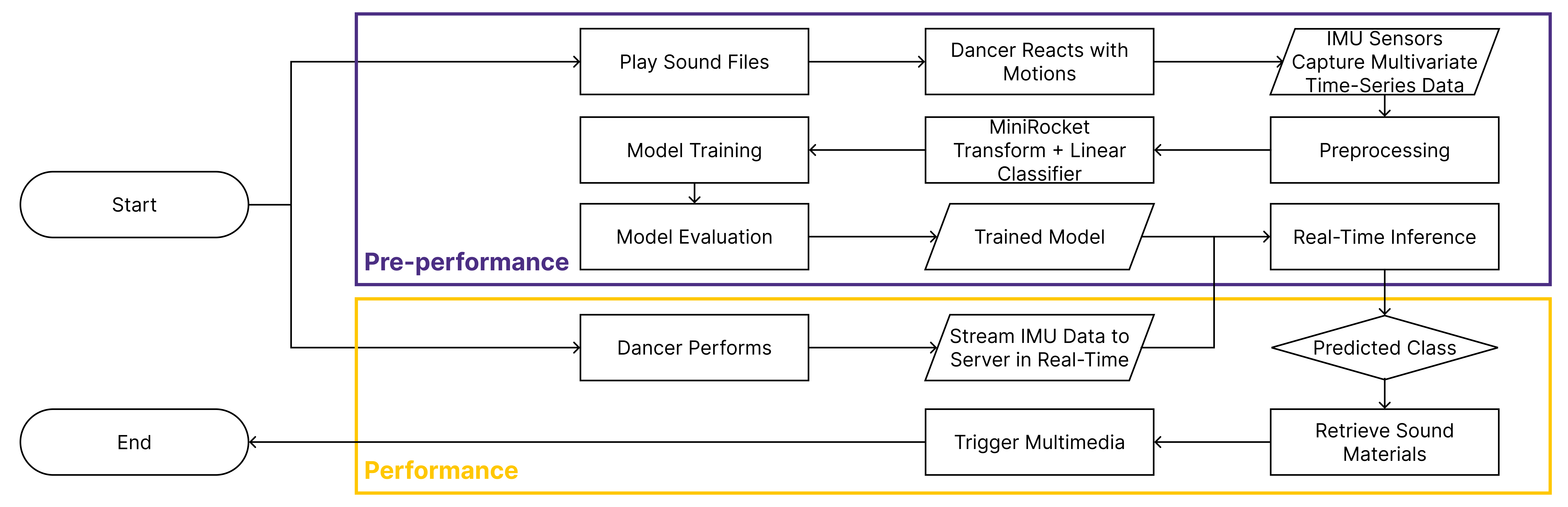} 
  \caption{\textbf{System pipeline for training and live application.} During the training stage (pre-performance), the dancer reacts to sound while wearing IMU sensors, which capture multivariate time-series motion data. This data is preprocessed and used to train a linear classifier with MiniRocket on a GPU server. During application (in performance), real-time data is streamed to the server, classified, and used to trigger corresponding multimedia elements, completing the loop from embodied memory to audiovisual output.}
  \label{fig:4.3.1}
\end{figure}

The system is not just conceptual but also technically robust for live use. Building on the human-centered mapping framework introduced in the previous section, we implement the technical pipeline below that establishes a closed feedback loop of human input and multimedia output. This process is illustrated in Figure \ref{fig:4.3.1}.

\begin{figure}[htbp]
  \centering
  \includegraphics[width=\textwidth]{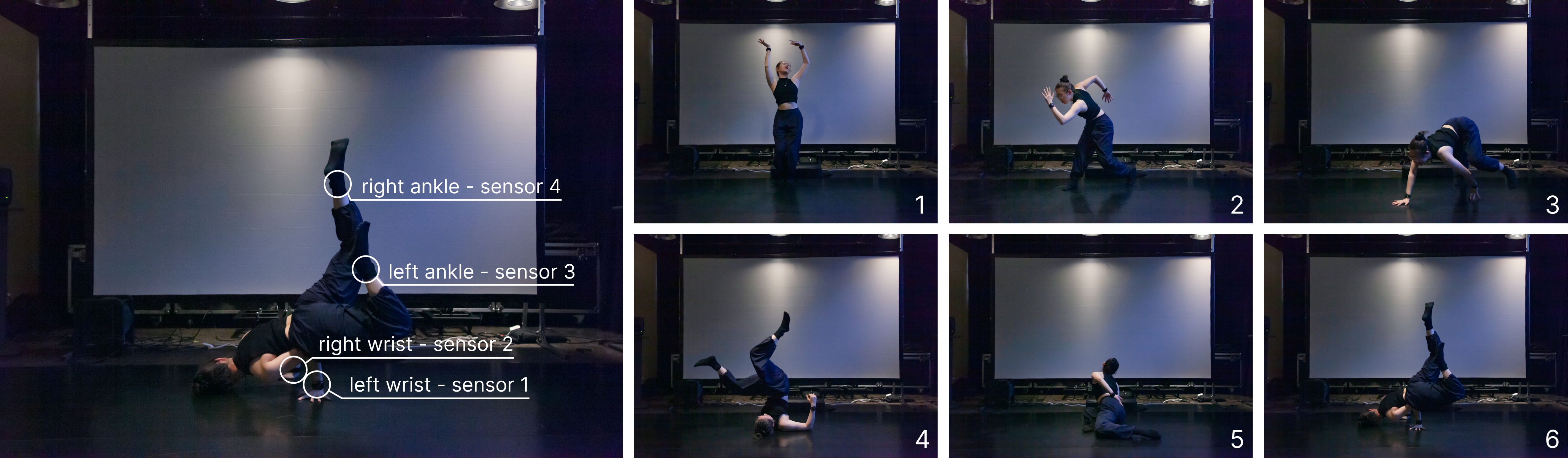}
  \caption{\textbf{Sensor placement and movement sequences documentation.} IMU sensors are affixed to both wrists and ankles (left panel). The right panel presents six frames from the recorded training session, each demonstrating distinct motion patterns corresponding to different sound stimuli. Movement class labels used in training are also marked in the images.}
  \label{fig:4.3.2}
\end{figure}

To collect movement data, as shown in Figure \ref{fig:4.3.2}, we use four wireless IMU sensors secured with adjustable fabric straps, each sensor with strap weighing approximately 0.025 kg (0.055 lb). The sensors are attached to the dancer’s wrists and ankles, transmitting real-time data via Bluetooth Low Energy (BLE). Each unit provides six channels (three-axis accelerometer and three-axis gyroscope), resulting in a total of 24 data channels. Sensor data is collected and processed at a sampling rate of 48 Hz.

While the dancer moves with the music, the IMU recordings are preprocessed into fixed-length chunks. We apply data augmentation techniques, including jittering and time warping, to increase the training dataset diversity. The preprocessed data features are extracted to train a ridge regression classifier which identifies data patterns and recognizes discrete motion types.

Once the model is trained, a Python script streams live IMU data from the dancer to our remote server, which performs real-time inference and returns predicted motion class labels along with their associated probabilities. The inference results are used to control sound and other media of the performance.

\section{Experiments and Results}

\begin{figure}[htbp]
  \centering
  \includegraphics[width=\textwidth]{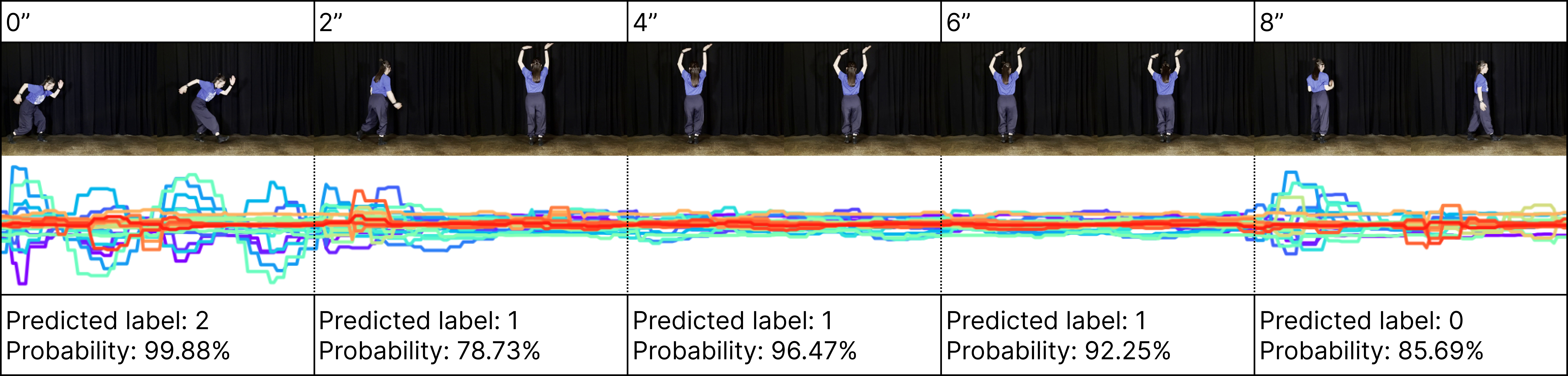}
  \caption{\textbf{Real-time motion classification during a 10‑second mock performance.} The dancer performs different movements, with the model inferring labels based on the dominant motion within each 2-second segment. During transitions, predicted probabilities tend to decrease, reflecting temporal ambiguity. Labels and probabilities are visualized alongside a time-aligned video strip and IMU signal plot.}
  \label{fig:5.1}
\end{figure}

We define seven distinct motion types as classification labels. To support continuous recognition, the multivariate IMU data is segmented into 2-second chunks and transmitted to the model for real-time inference. As illustrated in Figure \ref{fig:5.1} from a test run, the predicted motion class labels align closely with the dancer’s actual movements over time.

Since sample sizes vary across classes, we apply stratified 10-fold cross-validation to maintain balanced class distributions within each fold. The model achieves a mean accuracy of 96.05\% with a standard deviation of 2.89\%. The macro-averaged F1 score is 96.62\%. 

\begin{figure}[htbp]
  \centering
  \includegraphics[width=\textwidth]{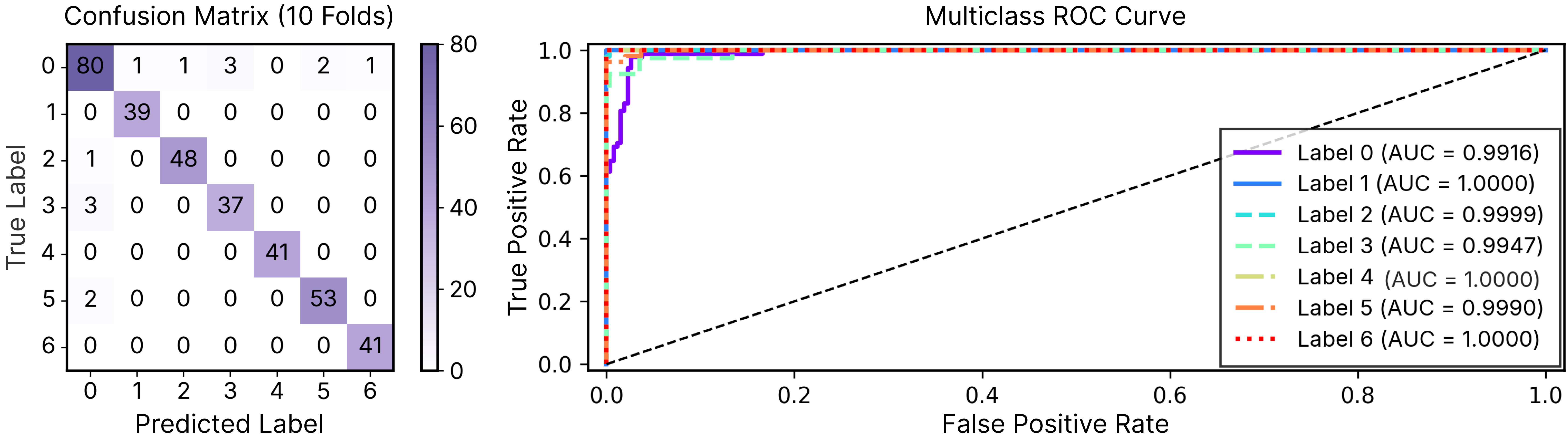}
  \caption{\textbf{Model performance evaluation.} A confusion matrix summed over 10-fold cross-validation (left panel) shows high classification accuracy across all motion classes. Class label 0 represents the negative class, and labels 1 to 6 refer to dance movements in Figure 2. The average multiclass ROC curve (right panel) for all seven labels is calculated, with all AUC scores above 0.99, indicating strong discriminability for each class.}
  \label{fig:5.2}
\end{figure}

Given the relatively small dataset of 648 samples, we further assess model reliability using a confusion matrix aggregated over 10 folds and a multiclass receiver operating characteristic (ROC) curve with area under the curve (AUC) scores reported for each class, as shown in Figure \ref{fig:5.2}.

The complete process, from streaming data to the remote server to receiving inference results in the Python script, takes less than 50 milliseconds. The inference itself takes approximately 15 milliseconds per data chunk. As a result, by integrating MiniRocket, this setup achieves efficient real-time motion classification which enables live control of the multimedia system during performance.

\begin{figure}[htbp]
  \centering
  \includegraphics[width=\textwidth]{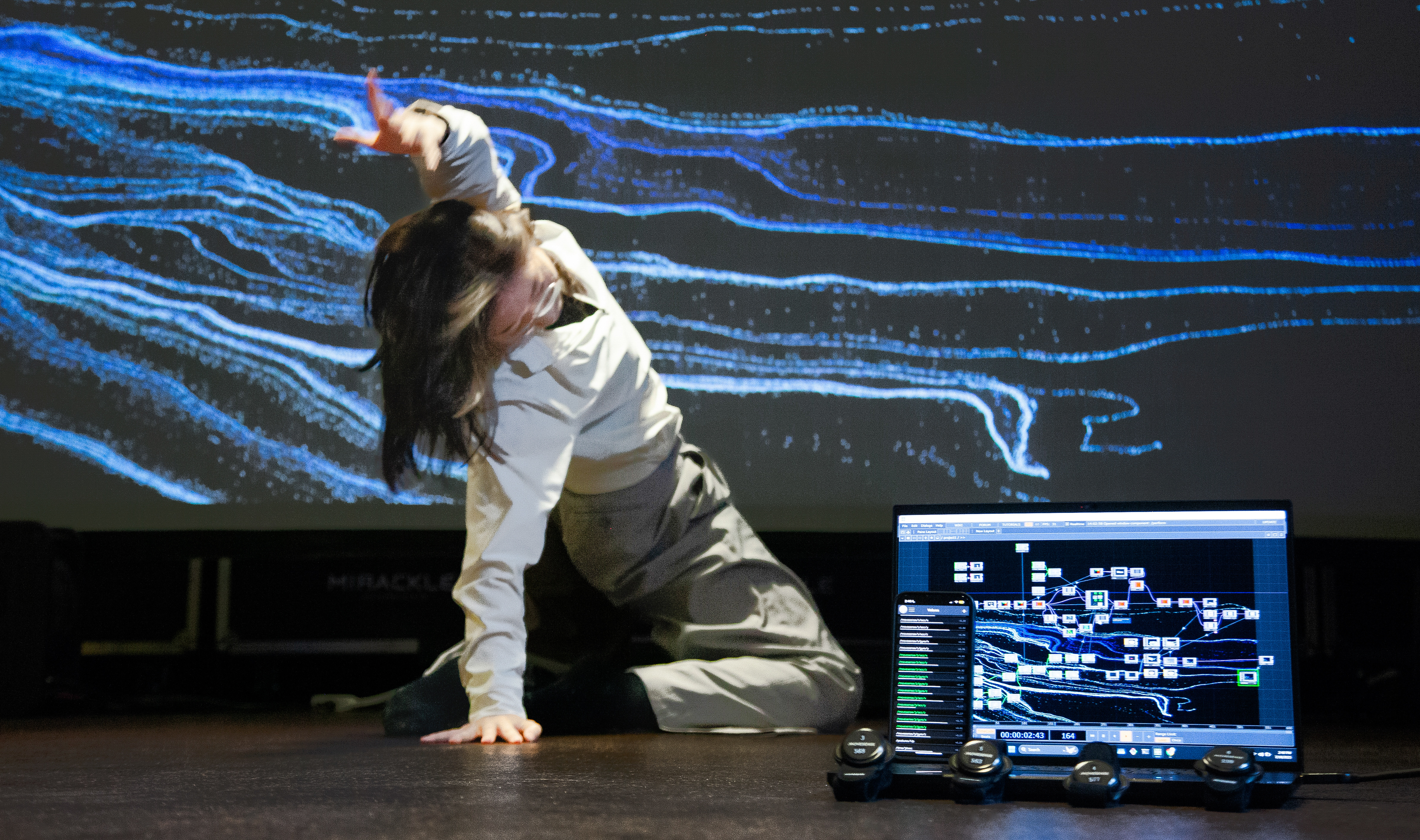}
  \caption{\textbf{Real-time human-machine interaction during rehearsal.} A dance artist performs in front of a projection controlled by our system described in Figure \ref{fig:4.3.1}. The laptop screen shows the multimedia interface, while wearable IMU sensors and a smartphone are used to stream and monitor real-time data. }
  \label{fig:5.3}
\end{figure}

In brief, the model is trained on a dataset tailored to the dancer’s movements and is optimized for instantaneous inference. A deeper analysis of the system applied in live performance will be conducted in future work. Figure \ref{fig:5.3} captures the system in use during a rehearsal session.

\section{Conclusion and Future Work}

Our project aims to reframe the human-machine relationship as contemporary AI systems increasingly shape creative practices. Through the ritual of co-performance, we ask: what happens when the machine remembers, rather than generates; when it listens, rather than speaks first? Our framework resists generalization and embraces subjectivity, centering bodily intuition and memory as a path to expressive, hybrid rituals of coexistence.

Future work will extend this project along several directions. First, we plan to expand the movement data archive by increasing both the size and diversity of motion types, strengthening the system’s vocabulary of embodied expression. Second, we aim to address motion transition ambiguity by developing higher‑granularity methods for more refined classification. Third, given the efficiency of MiniRocket combined with GPU acceleration, we seek to support real-time data collection of new motions and on-the-fly model retraining, evolving the system into an interactive machine learning tool. Fourth, we will build on our current use of sound‑mapping through memory‑based composition, exploring methods that deepen its expressive potential. Beyond technical optimization, we also envision applications in somatic education and therapeutic movement. Finally, we anticipate a live public performance, demonstrating this system as both a technical framework and an artistic collaborator.

In doing so, we continue to cultivate a dance-literate machine, which listens, remembers, and responds mindfully, opening up more dialogue between embodied human expression and computational perception with care.


\bibliographystyle{unsrt}
\bibliography{references}


\end{document}